\documentclass{article}
\usepackage{spconf,amsmath,graphicx}
\usepackage{color}

\title{End-to-End Monaural Multi-speaker ASR System without Pretraining}
\name{Xuankai Chang$^{1,2}$, Yanmin Qian$^{1}$, Kai Yu$^{1}$, Shinji Watanabe$^{2}$ \thanks{This work was down while Xuankai Chang was an intern at the Johns Hopkins University.}}
\address{
    $^1$SpeechLab, Department of Computer Science and Engineering, Shanghai Jiao Tong University, China\\
    $^2$Center for Language and Speech Processing, Johns Hopkins University, U.S.A\\
    xuank@sjtu.edu.cn, yanminqian@sjtu.edu.cn, kai.yu@sjtu.edu.cn, shinjiw@jhu.edu}
\begin{document}
\ninept
\maketitle
\begin{abstract}
Recently, end-to-end models have become a popular approach as an alternative to traditional hybrid models in automatic speech recognition (ASR). The multi-speaker speech separation and recognition task is a central task in cocktail party problem. In this paper, we present a state-of-the-art monaural multi-speaker end-to-end automatic speech recognition model. In contrast to previous studies on the monaural multi-speaker speech recognition, this end-to-end framework is trained to recognize multiple label sequences completely from scratch. The system only requires the speech mixture and corresponding label sequences, without needing any indeterminate supervisions obtained from non-mixture speech or corresponding labels/alignments. Moreover, we exploited using the individual attention module for each separated speaker and the scheduled sampling to further improve the performance. Finally, we evaluate the proposed model on the 2-speaker mixed speech generated from the WSJ corpus and the wsj0-2mix dataset, which is a speech separation and recognition benchmark. The experiments demonstrate that the proposed methods can improve the performance of the end-to-end model in separating the overlapping speech and recognizing the separated streams. From the results, the proposed model leads to $\sim10.0\%$ relative performance gains in terms of CER and WER respectively.
\end{abstract}
\section{Introduction}
\label{sec:intro}

In the deep learning era, single-speaker automatic speech recognition systems have achieved a lot of progress. Deep neural networks (DNN) and hidden markov model (HMM) based hybrid systems have attained surprisingly good performance \cite{DNN4ASR-hinton2012,DeepCNN-Sainath2013,HumanParity-Xiong2016}. Recently, there has been a growing interest in developing end-to-end models for speech recognition \cite{Joint-Kim2017,Hybrid-Watanabe2017,ESPnet-Watanabe2018}, in which the various modules of the hybrid systems, such as the acoustic model (AM) and language model (LM), are folded into a single neural network model. Two major approaches of end-to-end speech recognition systems are connectionist temporal classification \cite{Towards-Graves2014,EESEN-Miao2015,zhc00-chen-tasl17} and attention-based encoder-decoder \cite{End-Chorowski2014,Listen-Chan2016}. The performance of deep learning based conventional speech recognition systems has been reported to be comparable with, or even surpassing, human performance \cite{HumanParity-Xiong2016}. However, it is still extremely difficult to solve the cocktail party problem \cite{AMI-Carletta2005,Monaural-Cooke2010,Fifth-Barker2018,Past-Qian2018}, which refers to the task of separating and recognizing the speech from a specific speaker when it is interfered by noise and speech from other speakers.

To address the monaural multi-speaker speech separation and recognition problem, there has been a lot of research in single-channel multi-speaker speech separation and recognition, which aims to separate the overlapping speech and recognize the resulting separated speech individually, given a single-channel multiple-speaker mixtured speech.
In \cite{DeepClustering-Hershey2015,DeepClustering2-Isik2016}, a method called deep clustering (DPCL) was proposed for speech separation. DPCL separates the mixed speech by training a neural network to project each time-frequency (T-F) unit into a high-dimensional embedding space, in which pairs of T-F units are close to each other if they have the same dominating speaker and farther away otherwise. In addition to segmentation using k-means clustering, a permutation-free mask objective was proposed to refine the output \cite{DeepClustering2-Isik2016}. In \cite{Permutation-Yu2017,Multitalker-Kolbaek2017}, a speech separation method called permutation invariant training (PIT) was proposed to train a compact deep neural network with the objective that minimizes the average minimum square error of the best output-target assignment at the utterance level. PIT was later extended to train a speech recognition model for multi-speaker speech mixture by directly optimizing with the ASR objective \cite{Recognizing-Yu2017,Progressive-Chen2018,zhc00-chen-icassp18.2,Monaural-Chang2018,Single-Qian2018}. In \cite{End-Settle2018,End2End-Seki2018}, a joint CTC/attention-based encoder-decoder network for end-to-end speech recognition \cite{Joint-Kim2017,Hybrid-Watanabe2017} was applied to multi-speaker speech recognition. First, an encoder separates the mixed speech into hidden vector sequences for every speaker. Then an attention-based decoder is used to generate the label sequence for each speaker. To avoid label permutation problem, a CTC objective is used in permutation-free manner right after the encoder to determine the order of the label sequences. However, the model needs to first be pre-trained on single-speaker speech so that decent performance can be achieved.

In this paper, we explore several new methods to refine the end-to-end speech recognition model for multi-speaker speech. Firstly, we revise the model in \cite{End2End-Seki2018} so that pretraining on single-speaker speech is not required without loss of performance. Secondly, we propose to use speaker parallel attention modules. In previous work, the separated speech streams were treated equally in the decoder, regardless of the energy and speaker characteristics. We bring in multiple attention modules \cite{Attention-Vaswani2017} for each speaker to enhance the speaker tracing ability and to alleviate the burden of the encoder as well as \cite{Monaural-Chang2018}. Another method is to use scheduled sampling \cite{Scheduled-Bengio2015} to randomly choose the token from either the ground truth or the model prediction as the history information, which reduces the gap between training and inference in the sequence prediction tasks. This would be extremely helpful in our setup, since the separation is not always perfect and we often observe mixed label results. Schedule sampling can help to recover such errors during inference.

The rest of the paper is organized as follows: In Section \ref{sec:multi-speakerE2E}, the end-to-end monaural multi-speaker ASR model and the proposed new methods are described. In Section \ref{sec:experiment}, we evaluate the proposed approach on the 2-speaker mixing WSJ data set, and the experiments and analysis are given. Finally the paper is concluded in Section \ref{sec:conclusion}.

\section{End-to-End Multi-speaker Joint CTC/Attention-based Encoder-Decoder}
\label{sec:multi-speakerE2E}
In this section, we first describe the end-to-end ASR system for multi-speaker speech that has been used in \cite{End2End-Seki2018}. Then we introduce two techniques to improve the training process and performance of the end-to-end ASR multi-speaker system, namely the speaker parallel attention and scheduled sampling \cite{Scheduled-Bengio2015}.

\subsection{End-to-End Multi-speaker ASR}
\label{ssec:multi-speakerE2E}

In \cite{Joint-Kim2017,Hybrid-Watanabe2017,Joint-Hori2017}, an end-to-end speech recognition model was proposed to take advantage of both the Connectionist Temporal Classification (CTC) and attention-based encoder-decoder, in aim of using the CTC to enhance the alignment ability of the model. An end-to-end model for multi-speaker speech recognition was brought up in \cite{End2End-Seki2018}, extending the joint CTC/attention-based encoder-decoder network to be applied on multi-speaker speech mixtures and to allow the permutation-free training in the objective function to address the permutation problem. The model is shown in Fig.\ref{fig:multi-spkrE2E}, in which the modules \textit{Attention 1} and \textit{Attention 2} share parameters. The input speech mixture is first explicitly separated into multiple sequences of vectors in the encoder, each representing a speaker source. These sequences are fed into the decoder to compute the conditional probabilities.

The encoder of the model can be divided into three stages, namely the $\mathrm{Encoder_{Mix}}$, $\mathrm{Encoder_{SD}}$ and $\mathrm{Encoder_{Rec}}$. Let $\mathbf{O}$ denote an input speech mixture from $S$ speakers. The first stage, $\mathrm{Encoder_{Mix}}$, is the mixture encoder, which encodes the input speech mixture $\mathbf{O}$ as an intermediate representation $\mathbf{H}$. Then, the representation $\mathbf{H}$ is processed by $S$ speaker-different (SD) encoders, $\mathrm{Encoder_{SD}}$, with the outputs being referred to as feature sequences $\mathbf{H}^s, s=1,\cdots,S$. $\mathrm{Encoder_{Rec}}$, the last stage, transforms the features sequences to high-level representations $\mathbf{G}^s, s=1,\cdots,S$. The encoder is computed as
\begin{align}
    \mathbf{H}   &= \mathrm{Encoder_{Mix}} (\mathbf{O}) \\
    \mathbf{H}^s &= \mathrm{Encoder_{SD}}^s (\mathbf{H}), s=1, \cdots, S \\
    \mathbf{G}^s &= \mathrm{Encoder_{Rec}} (\mathbf{H}^s), s=1, \cdots, S
\end{align}

In the single-speaker joint CTC/attention-based encoder-decoder network, the CTC objective function is used to train the attention model encoder as an auxiliary task right after the encoder \cite{Joint-Kim2017,Hybrid-Watanabe2017,Joint-Hori2017}. While in the multi-speaker framework, the CTC objective function is also used to perform the permutition-free training as in Eq.\ref{eq:perm}, which is referred to as permutation invariant training in \cite{Past-Qian2018,Permutation-Yu2017,Recognizing-Yu2017,Progressive-Chen2018,zhc00-chen-icassp18.2,Monaural-Chang2018,Single-Qian2018,Adaptive-Chang2018,Knowledge-Tan2018}.
\begin{align}
    \hat{\pi} = \arg \min_{\pi \in \mathcal{P}} \sum_s \mathrm{Loss_{ctc}} (\mathbf{Y}^s, \mathbf{R}^{\pi(s)}), \label{eq:perm}
\end{align}
where $\mathbf{Y}^s$ is the output sequence variable computed from the encoder output $\mathbf{G}^s$, $\pi(s)$ is the $s$-th element in a permutation $\pi$ of $\{1, \cdots, S\}$, and $\mathbf{R}$ is the reference labels for $S$ speakers. Later, the permutation $\hat{\pi}$ with minimum CTC loss is used for the reference labels in the attention-based decoder in order to reduce the computational cost.

After obtaining the representations $\mathbf{G}^s, s=1,\cdots,S$ from the encoder, an attention-based decoder network is used to decode these streams and output label sequence $\mathbf{Y}^s$ for each representation stream according to the permutation determined by the CTC objective function. For each pair of representation and reference label index $(s, \hat{\pi}(s))$, the decoding process is described as the following equations:
\begin{align}
    p_{\mathrm{att}}(Y^{s, \hat{\pi}(s)} | \mathbf{O}) &= \prod_{n} p_{\mathrm{att}}(y^{s, \hat{\pi}(s)}_n | \mathbf{O}, y^{s, \hat{\pi}(s)}_{1:n-1}) \label{eq:seq_prob} \\
    c^{s, \hat{\pi}(s)}_n &= \mathrm{Attention}(a^{s, \hat{\pi}(s)}_{n-1}, e^{s, \hat{\pi}(s)}_{n-1}, \mathbf{G}^s) \label{eq:attention} \\
    e^{s, \hat{\pi}(s)}_n &= \mathrm{Update}(e^{s, \hat{\pi}(s)}_{n-1}, c^{s, \hat{\pi}(s)}_{n-1}, y^{\hat{\pi}(s)}_{n-1}) \label{eq:dec-hid} \\
    y^{s, \hat{\pi}(s)}_n &\sim \mathrm{Decoder}(c^{s, \hat{\pi}(s)}_n, y^{\hat{\pi}(s)}_{n-1}) \label{eq:dec-out}
\end{align}
where $c^{s, \hat{\pi}(s)}_n$ denotes the context vector, $e^{s, \hat{\pi}(s)}_n$ is the hidden state of the decoder, and $r^{\hat{\pi}(s)}_{n}$ is the $n$-th element in the reference label sequence. During training, the reference label $r^{\hat{\pi}(s)}_{n-1}$ in $\mathbf{R}$ is used as a history  in the manner of teacher-forcing, instead of $y^{\hat{\pi}(s)}_{n-1}$ in Eq.\ref{eq:dec-hid} and Eq.\ref{eq:dec-out}.
And, Eq.\ref{eq:seq_prob} means the probability of the target label sequence $\mathbf{Y}=\{y_1,\cdots,y_N\}$ that the attention-based encoder-decoder predicted, in which the probability of $y_n$ at $n$-th time step is dependent on the previous sequence $y_{1:n-1}$.

The final loss function is defined as
\begin{align}
    \mathcal{L}_{\mathrm{mtl}} &= \lambda \mathcal{L}_{\mathrm{ctc}} + (1-\lambda) \mathcal{L}_{\mathrm{att}}, \label{eq:loss_mtl} \\
    \mathcal{L}_{\mathrm{ctc}} &= \sum_s Loss_{\mathrm{ctc}} (\mathbf{Y}^s, \mathbf{R}^{\hat{\pi}(s)}), \\
    \mathcal{L}_{\mathrm{att}} &= \sum_s Loss_{\mathrm{att}} (\mathbf{Y}^{s,\hat{\pi}(s)}, \mathbf{R}^{\hat{\pi}(s)}),
\end{align}
where $\lambda$ is the interpolation factor, and $0 \leq \lambda \leq 1$.

\subsection{Speaker parallel attention modules}
\label{ssec:multi-att}

\begin{figure}[htb]
\begin{minipage}[b]{1.0\linewidth}
  \centering
  \centerline{\includegraphics[width=8.5cm]{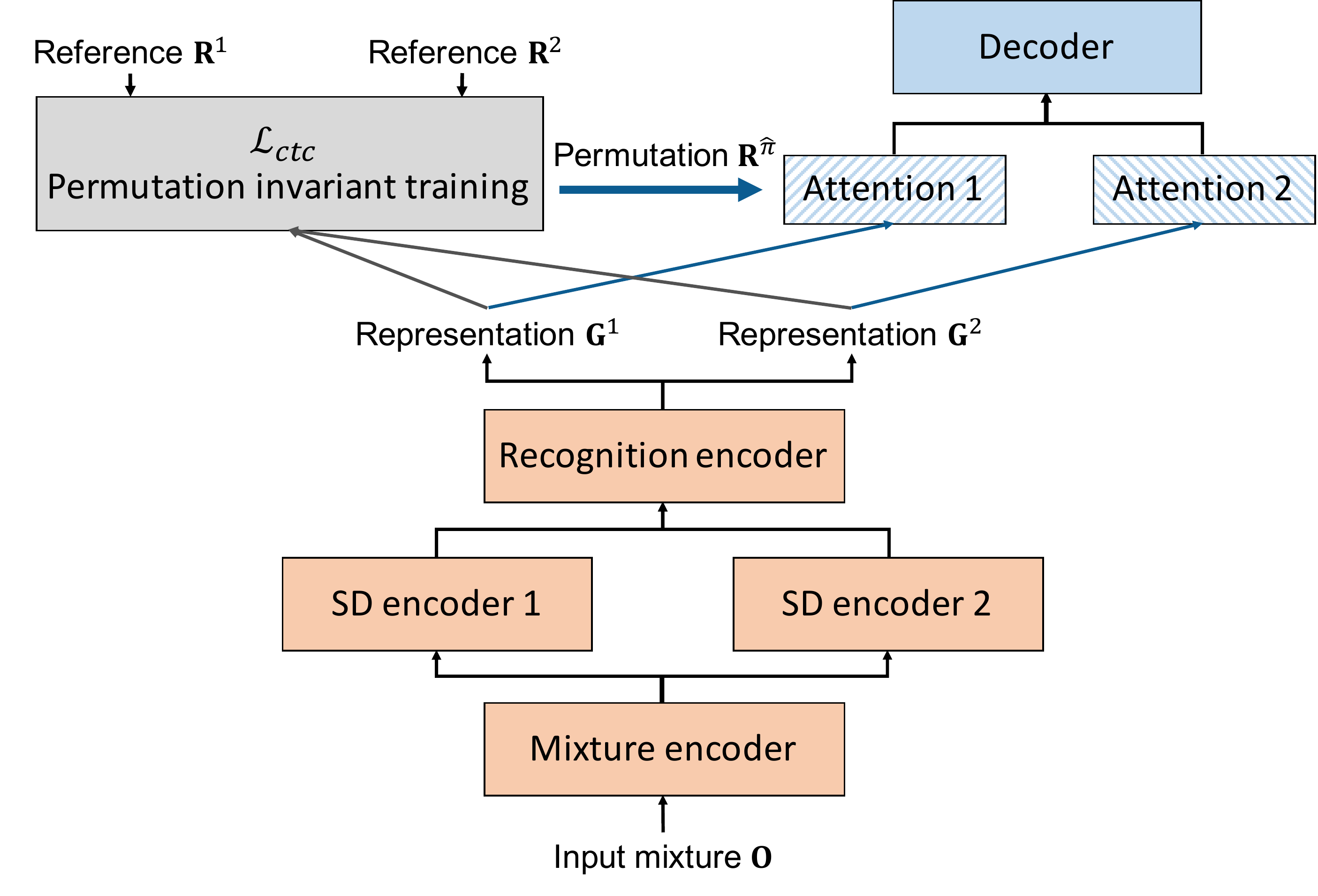}}
  \caption{End-to-End Multi-speaker Speech Recognition Model in the 2-Speaker Case}
  \label{fig:multi-spkrE2E}
\end{minipage}
\end{figure}
Due to the differences in the characteristics of speakers and energy, the encoder usually has to compensate for those differences while separating the speech. The motivation of speaker parallel attention module that we proposed is to alleviate the burden for the encoder and to make the attention-decoder learn to filter the separated speech as well while keeping the model compact. In light of \cite{Monaural-Chang2018}, we proposed to use independent attention modules called speaker parallel attention. Fig.\ref{fig:multi-spkrE2E} illustrates the architecture of the model, in which \textit{Attention 1} and \textit{Attention 2} are not sharing. The computation process in Eq.\ref{eq:attention} should be rewritten in a stream-specific way, in particular for the $s$-th stream, as:
\begin{align}
    c^{s, \hat{\pi}(s)}_n, a^{s, \hat{\pi}(s)}_n &= \mathrm{Attention}^s(a^{s, \hat{\pi}(s)}_{n-1}, c^{s, \hat{\pi}(s)}_{n-1}, \mathbf{G}^s) \label{eq:new_attention}
\end{align}

\subsection{Scheduled sampling}
\label{ssec:sampling}
We generally trained the decoder network in a teacher-forcing fashion, which means the reference label token $r_n$, not the predicted token $y_n$, is used to predict the next token in the sequence during training. However, during inference, we are only accessible to the predicted token $y_n$ from the model itself. This difference may lead to performance degradation, especially in the multi-speaker speech recognition task susceptible to the label permutation problem. We alleviate this problem by using the scheduled sampling technique \cite{Scheduled-Bengio2015}. During training, whether the history information is chosen from the ground truth label or the prediction is done randomly with a probability of $p$ from the the prediction and $(1-p)$ from ground truth. Thus Eq.\ref{eq:dec-hid} and Eq.\ref{eq:dec-out} should be changed as:
\begin{align}
    e^{s, \hat{\pi}(s)}_n &= \mathrm{Update}(e^{s, \hat{\pi}(s)}_{n-1}, c^{s, \hat{\pi}(s)}_{n-1}, h), \\
    y^{s, \hat{\pi}(s)}_n &\sim \mathrm{Decoder}(c^{s, \hat{\pi}(s)}_n, h),
\end{align}
where
\begin{align}
    b &\sim Bernoulli(p), \label{eq:bernoulli} \\
    h &= \left\{\begin{matrix}
    r^{\hat{\pi}(s)}_{n-1}, & if\ b=0 \\
    y^{\hat{\pi}(s)}_{n-1}, & if\ b=1
    \end{matrix}\right.
\end{align}

\section{Experiment}
\label{sec:experiment}

\subsection{Experimental setup}
\label{ssec:setup}
To evaluate our method, we used the artificially generated single-channel two-speaker mixed signals from the Wall Street Journal (WSJ) speech corpus according to \cite{End2End-Seki2018}, using the tool released by MERL\footnote{http://www.merl.com/demos/deep-clustering/create-speaker-mixtures.zip}. We used the WSJ SI284 to generate the training data, Dev93 for development and Eval92 for evaluation. The durations for the training, development and evaluation sets of the mixed data are 98.5 hr, 1.3 hr, and 0.8 hr respectively. In section \ref{ssec:wsj0-2mix-exp}, we also compared our model with previous works on the wsj0-2mix dataset, which is a standard speech separation and recognition benchmark \cite{DeepClustering-Hershey2015,DeepClustering2-Isik2016,End-Settle2018}.

The input feature is 80-dimensional log Mel filterbank coefficients with pitch features and their delta and delta delta features extracted using the Kaldi \cite{Kaldi-Povey2011}. Zero mean and unit variance are used to normalize the input features. All the joint CTC/attention-based encoder-decoder networks for end-to-end speech recognition were built based on the ESPnet \cite{ESPnet-Watanabe2018} framework. The networks were initialized randomly from uniform distribution in the range $-0.1$ to $0.1$. We used the AdaDelta algorithm with $\rho=0.95$ and $\epsilon=1e-8$. During training, we set the interpolation factor $\lambda$ in Eq.\ref{eq:loss_mtl} to be $0.2$. We revise the deep neural network, replacing the original encoder layers with shallower but wider layers \cite{Improved-Zeyer2018}, so that the performance can be good enough without pre-training on single-speaker speech.

To make the model comparable, we set all the neural network models to have the same depth and similar size. We use the VGG-motivated CNN layers and bidirectional long-short term memory recurrent neural networks with projection (BLSTMP) as the encoder. The total depth of the encoder is 5, namely two CNN blocks and three layer BLSTMP layers. For all models, the decoder network has 1 layer of unidirectional long-short term memory network (LSTM) with 300 cells.

During decoding, we combined both the joint CTC/attention score and the pretrained word-level recurrent neural network language model (RNNLM) score, which had 1-layer LSTM with 1000 cells and was trained on the transcriptions from WSJ SI284, in a shallow fusion manner. We set the beam width to be 30. The interpolation factor $\lambda$ we used during decoding was $0.3$, and the weight for RNNLM was $1.0$.

\subsection{Performance of baseline systems}
\label{ssec:baseline}
In this section, we describe the performance of the baseline E2E ASR systems on multi-speaker mixed speech. The first baseline system is the joint CTC/attention-based encoder-decoder network for single-speaker speech trained on WSJ corpus, whose performance is $0.9$\% in terms of CER and $1.9$\% in terms of WER on the eval92\_5k test set with the closed vocabulary. In the encoder, there are 3 layers of BLSTMP following the CNN and each BLSTMP layer has 1024 memory cells in each direction. The second baseline system is the joint CTC/attention-based encoder-decoder network for multi-speaker speech. The 2-layer CNN is used as the $\mathrm{Encoder_{Mix}}$. The depth of the following BLSTMP layers is also 3 including 1 layer of BLSTMP as the $\mathrm{Encoder_{SD}}$ and 2 layers of BLSTMP as the $\mathrm{Encoder_{Rec}}$. The attention-decoder in the multi-speaker system is shared among representations $\mathbf{G}^s$, which is of the same architecture with single-speaker system. The results are shown in Table \ref{tab:baseline}.
\begin{table}[th]
    \centering
    \begin{tabular}{c|c|c}
    \hline
    Model & dev CER & eval CER \\
    \hline
    single-speaker & 79.13 & 76.52 \\
    \hline
    multi-speaker \cite{End2End-Seki2018} & n/a & 13.7 \\
    multi-speaker & 15.14 & 12.20 \\
    \hline
    \hline
    Model & dev WER & eval WER \\
    \hline
    single-speaker & 113.47 & 112.21 \\
    \hline
    multi-speaker & 24.90 & 20.43 \\
    \hline
    \end{tabular}
    \vspace{-1ex}
    \caption{Performance (Avg. CER \& WER) (\%) on 2-speaker mixed WSJ corpus. Comparison between End-to-End single-speaker and multi-speaker joint CTC/attention-based encoder-decoder systems}
    \label{tab:baseline}
\end{table}

In the case of single-speaker, the CER and WER is measured by comparing the output against the reference labels of both speakers. From the table, we can see that the speech recognition system designed for multi-speaker can improve the performance for the overlapped speech significantly, leading to more than $80.0\%$ relative reduction on both average CER and WER. As a comparison, we also include the CER result from \cite{End2End-Seki2018} in the table, and it shows that the newly constructed end-to-end multi-speaker system without pretraining in this work can achieve better performance.

\subsection{Performance of speaker parallel attention with scheduled sampling}
\label{ssec:newmodel}
In this section we report the results of the evaluation of our proposed methods. The first method is the speaker parallel attention, introducing independent attention modules for each speaker source instead of using a shared attention module. The rest of the network is kept the same as the baseline multi-speaker model, containing a 2-layer CNN $\mathrm{Encoder_{Mix}}$, 1-layer BLSTMP $\mathrm{Encoder_{SD}}$, a 2-layer BLSTMP $\mathrm{Encoder_{Rec}}$, and a shared 1-layer LSTM as the decoder network. The performance is illustrated in the Table \ref{tab:proposed}. The speaker parallel attention module reduces the average CER by $9\%$ and average WER by $8\%$ relatively. From the results we can tell that the CER is high, so the gap is large between the training and inference using the teacher-forcing fashion. Thus we adopted the scheduled sampling method with probability $p=0.2$ in Eq. \ref{eq:bernoulli}, which lead to a further improvement in performance. Finally, the system using both speaker parallel attention and scheduled sampling can obtain relative $\sim10.0\%$ reduction on both CER and WER on the evaluation set.
\begin{table}[th]
    \centering
    \begin{tabular}{l|c|c}
    \hline
    Model & dev CER & eval CER \\
    \hline
    multi-speaker (baseline) & 15.14 & 12.20 \\
    \hline
    + speaker parallel attention & 14.80 & 11.11 \\
    \ \ ++ scheduled sampling & \textbf{14.78} & \textbf{10.93} \\
    \hline
    \hline
    Model & dev WER & eval WER \\
    \hline
    multi-speaker (baseline) & 24.90 & 20.43 \\
    \hline
    + speaker parallel attention & 24.88 & 18.76 \\
    \ \ ++ scheduled sampling & \textbf{24.52} & \textbf{18.44} \\
    \hline
    \end{tabular}
    \vspace{-1ex}
    \caption{Performance (Avg. CER \& WER) (\%) on 2-speaker mixed WSJ corpus. Comparison between End-to-End multi-speaker joint CTC/attention-based encoder-decoder systems}
    \label{tab:proposed}
    \vspace{-2ex}
\end{table}

\begin{figure}[htb]
\begin{minipage}[b]{\linewidth}
  \centering
  \includegraphics[width=\linewidth]{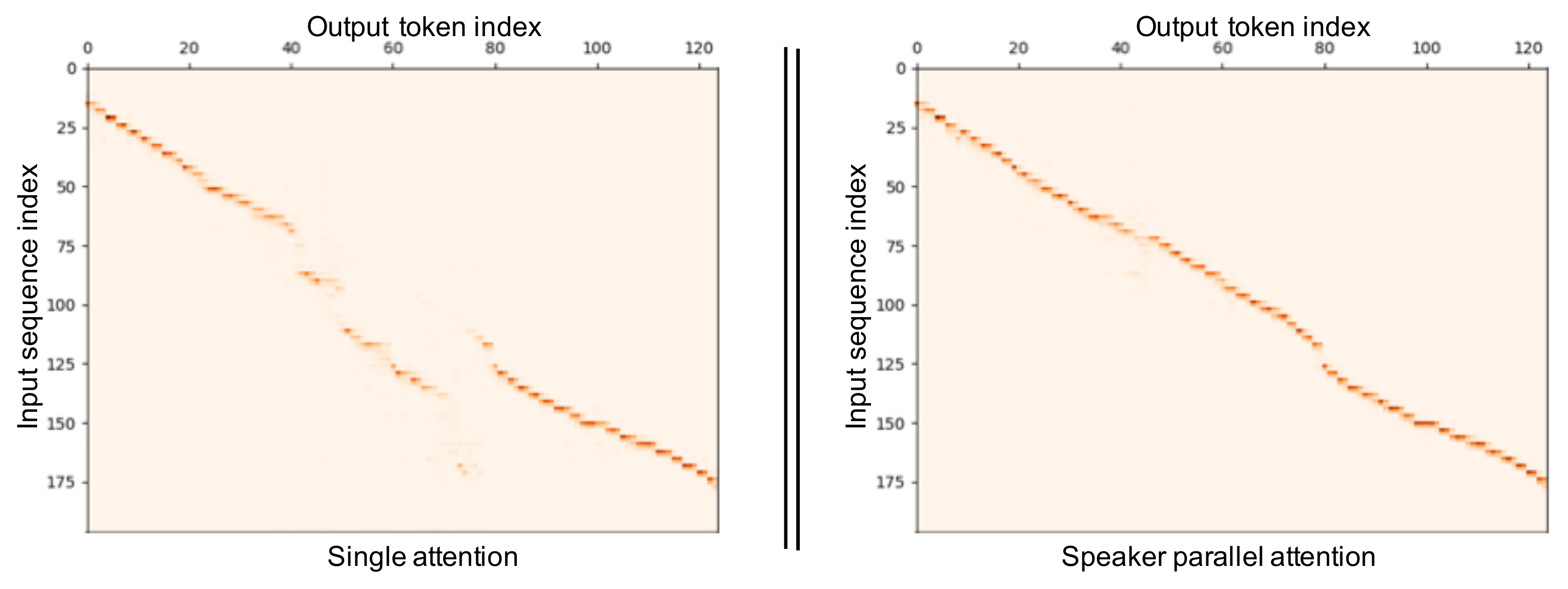}
  \centerline{(a) Attention weights for speaker 1}
\end{minipage}
\begin{minipage}[b]{\linewidth}
  \centering
  \includegraphics[width=\linewidth]{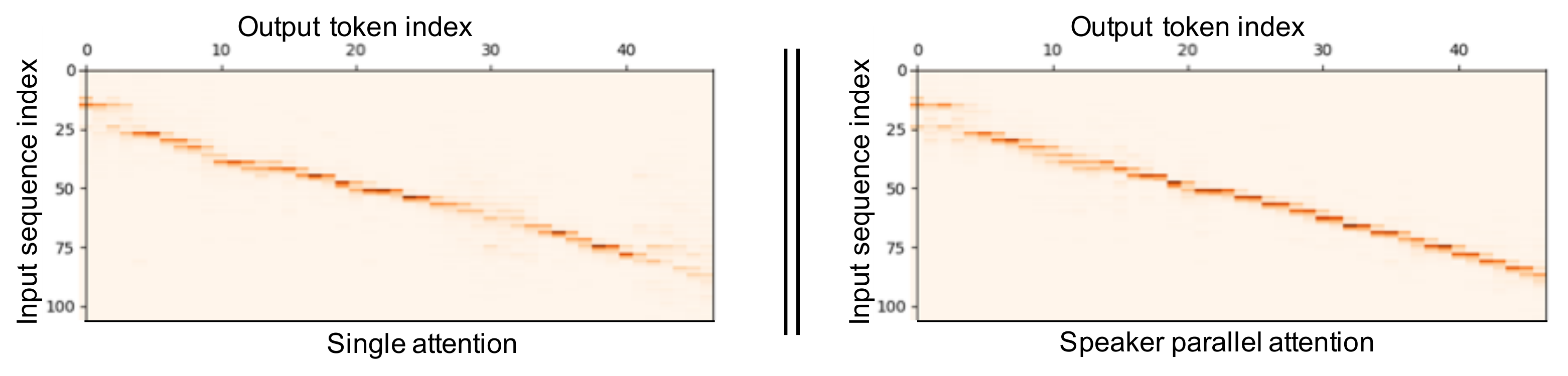}
  \centerline{(b) Attention weights for speaker 2}
\end{minipage}
\vspace{-2ex}
\caption{Visualization of the attention weights sequences for two overlapped speakers. The left part is from the previous single-attention multi-speaker end-to-end model and the right part is from the proposed speaker-parallel-attention multi-speaker end-to-end model.}
\label{fig:att-ws}
\vspace{-2ex}
\end{figure}
We show the visualization of the attention weights sequences for two overlapped speakers, generated by the baseline single-attention multi-speaker end-to-end model and the proposed speaker-parallel-attention multi-speaker end-to-end model individually. The horizontal axis represents the output token sequence and the vertical axis represents the input sequence to the attention module. The left parts of Figures.\ref{fig:att-ws} (a) and (b) show the attention weights for speaker 1 and speaker 2 generated by the previous single-attention model. The right parts show the attention weights generated by the proposed speaker-parallel-attention model. We can observe that the right parts are more smooth and clear, and the attention weights are more concentrated. This observation conforms with the characteristics of alignments between output sequence and input sequence for speech recognition, and further shows the superiority of the proposed speaker parallel attentions.

\subsection{Comparison with previous work}
\label{ssec:wsj0-2mix-exp}
We then compared our work with other related works. We trained and tested our model on wsj0-2mix dataset that was first used in \cite{DeepClustering-Hershey2015}. Table \ref{tab:relatedworks} shows the WER results of hybrid systems including PIT-ASR \cite{Single-Qian2018}, DPCL-based speech separation with Kaldi-based ASR \cite{DeepClustering2-Isik2016}, and the end-to-end systems constructed in \cite{End2End-Seki2018} and ours in this paper.
These were evaluated under the same evaluation data and metric as in \cite{DeepClustering2-Isik2016} based on the wsj0-2mix. Noted that the model in \cite{End2End-Seki2018} was trained on a different, larger training dataset than that used in other experiments. From Table. \ref{tab:relatedworks}, we can observe that our new system constructed by the proposed methods in this paper is significantly better than the others.
\begin{table}[th]
    \centering
    \begin{tabular}{c|c}
    \hline
    Model & Avg. WER \\
    \hline
    DPCL+ASR \cite{DeepClustering2-Isik2016} & 30.8 \\
    PIT-ASR \cite{Single-Qian2018} & 28.2 \\
    End-to-end ASR (Char/Word-LM) \cite{End2End-Seki2018} & 28.2 \\
    Proposed End-to-end ASR with SPA (Word LM) & \textbf{25.4} \\
    \hline
    \end{tabular}
    \vspace{-1ex}
    \caption{WER (\%) on 2-speaker mixed \textbf{WSJ0} corpus. The comparison is done between our proposed end-to-end ASR with speaker parallel attention (SPA) and previous works including DPCL+ASR, PIT-ASR and end-to-end ASR systems.}
    \label{tab:relatedworks}
    \vspace{-2ex}
\end{table}

\section{Conclusion}
\label{sec:conclusion}

In this paper, we have introduced a state-of-the-art end-to-end multi-speaker speech recognition system under the joint CTC/attentin-based encoder-decoder framework. More specifically, a new neural network architecture enabled us to train the model from random initialization. And we adopted the speaker parallel attention module and scheduled sampling to improve performance over the previous end-to-end multi-speaker speech recognition system. The experiments on the 2-speaker mixed speech recognition show that the proposed new strategy can obtain a relative $\sim10.0\%$ improvement on CER and WER reduction.

\section{Acknowledgement}
We are also grateful to Matthew Maciejewski and Tian Tan for their comments on an earlier version of the manuscript.


%
%
%


\bibliographystyle{IEEEbib}
\bibliography{strings,refs}

\end{document}